\documentclass[11pt]{article}
\usepackage{nodalida2017}
\usepackage{mathptmx}
\usepackage{url}
\usepackage{latexsym}
\usepackage{todonotes}

\usepackage{multirow}
\usepackage{booktabs}
\usepackage{amssymb}
\usepackage{supertabular}

\usepackage{tikz-dependency}
\depstyle{depstd}{arc edge,arc angle=50,text only label,label style=above}

\def\arraystretch{0.8}

\newcommand{\depright}[2]{#1 $\curvearrowright$ #2}
\newcommand{\depleft}[2]{#1 $\curvearrowleft$ #2}
\newcommand{\dep}[1]{\texttt{#1}}

\title{A Systematic Comparison of Syntactic Representations of \\Dependency Parsing}

\author{Guillaume Wisniewski \\
  LIMSI, CNRS, Univ. Paris-Sud \\ Universit\'{e} Paris-Saclay \\ 91\,405 Orsay, France \\
  {\tt guillaume.wisniewski@limsi.fr} \\\And
  Oph\'elie Lacroix \\
  DIKU, University of Copenhagen\\
  University Park 5 \\
  2100 Copenhagen \\
  {\tt lacroix@di.ku.dk} \\}

\date{}

\begin{document}
\maketitle
\begin{abstract}
  We compare the performance of a transition-based parser in regards
  to different annotation schemes. We propose to convert some specific
  syntactic constructions observed in the universal dependency
  treebanks into a so-called more standard representation and to
  evaluate parsing performances over all the languages of the
  project. We show that the ``standard'' constructions do not lead
  systematically to better parsing performance and that the scores
  vary considerably according to the languages.
\end{abstract}

\section{Introduction}

Many treebanks have been developed for dependency
parsing, following different annotations conventions. The divergence
between the guidelines can results from both the
theoretical linguistic principles governing the choices of head status
and dependency inventories or to improve the performance of
down-stream applications~\cite{elming13downstream}. Therefore it is
difficult to compare parsing performance across languages or even
across the different corpora of a single language. 

Two projects of unified treebanks have recently emerged: the
HamleDT~\cite{zeman2014hamledt} and the Universal Dependency Treebank
(UDT)~\cite{Mcdonald13universal}. They aim at harmonizing annotation
schemes (at the level of PoS-tags and dependencies) between languages
by converting existing treebanks to the new scheme.  These works have
led to the creation of the Universal Dependencies (UD) project
\cite{universaldependencies13} that gathers treebanks for more than 45
languages (v1.3).

The UD annotation scheme has been designed to facilitate the transfer
of annotations across languages: similar syntactic relations are
represented by similar syntactic structures in different languages,
and relations tend to hold between content words rather than through
function words.  However, \cite{schwartz2012learnability} showed that,
for English, some of the choices made to increase the sharing of
structures between languages actually hurts parsing performance. Since
then the UD scheme has been hypothesized to be sub-optimal for
(monolingual) parsing.



In this work, we propose to systematically compare the parsing
performance of alternative syntactic representations over all the
languages of the UD project. We design a set of rules\footnote{Source
  code to transform between the various dependency structures we
  consider can be downloaded from
  \url{https://perso.limsi.fr/wisniews/recherche/#dependency-transformations}}
to automatically modify the representation of several syntactic
constructions of the UD to alternative representations proposed in the
literature (\textsection~\ref{sec:transfo}) and evaluate whether these
transformations improve parsing performance or
not~(\textsection~\ref{sec:experiments}). Further we try to relate the
choice of the syntactic representation to different measure of
\emph{learnability} to see if it is possible to predict which
representation will achieve the best parsing performance.


\section{Related Work} \label{sec:relatedwork}

Since \cite{nilsson06graph} many works have shown that well-chosen
transformations of syntactic representations can greatly improve the
parsing accuracy achieved by dependency parsers.
\cite{schwartz2012learnability} shows that ``selecting one
representation over another may affect parsing performance''.
Focusing on English, they compare parsing performance through
several alternatives 
and conclude that
parsers prefer attachment via function word over content-word
attachments. 
They argue that the \emph{learnability} of a
representation, estimated by the accuracy within this representation
is a good criterion for selecting a syntactic representation among
alternatives.

More recently, \cite{delhoneux2016representation} studies the
representation of verbal constructions to see if parsing works better
when auxiliaries are the head of auxiliary dependency relations, which
is not the case in UD. They highlight that the parsing benefits from
the disambiguation of PoS tags for main verbs and auxiliaries in UD
PoS tagset even if the overall parsing accuracy decreases.

To the best of our knowledge, \cite{rosa2015multi} is the only work to
study the impact of the annotation scheme on the performance of
transferred parsers. It compares the Prague annotation style used in
the HamleDT~\cite{zeman2014hamledt} with the Stanford
style~\cite{demarneffe2008stanford} that has inspired the UD
guidelines and shows that Prague style results in better parsing
performance. Nevertheless --- with a particular focus on the
adposition attachment case --- the Stanford style is advantageous for
delexicalized parsing transfer.

Finally, \cite{silveira15does} performs an analysis very similar to
ours and find that, for English, UD is a good parsing
representation. More recently, \cite{kohita17multilingual} shows that
it is possible to improve parsing performance for a wide array of
language by converting the dependency structure back-and-forth.



\def\arraystretch{1}
\begin{table*}[ht!]
\centering
\begin{small}
\begin{tabular}{|p{60pt}|l|c|c|}
\hline
\multicolumn{2}{|c|}{\textbf{Syntactic Functions}} & \multicolumn{2}{|c|}{\textbf{Annotation Scheme}} \\
& \textbf{UD relations} & \textbf{UD} & \textbf{Alternative} \\
\hline
	Clause 	 
	&
	\dep{mark}
	& 
	\multirow{2}{*}{
	\begin{dependency}[depstd]
		\begin{deptext}[column sep=0.03cm]
		to \& read  \\
		\end{deptext}
	    \depedge[edge start x offset=3pt]{2}{1}{}
	\end{dependency}
	}
	&
	\multirow{2}{*}{
	\begin{dependency}[depstd]
		\begin{deptext}[column sep=0.03cm]
		to \& read  \\
		\end{deptext}
	    \depedge[edge start x offset=-4pt]{1}{2}{}
	\end{dependency} 
	}
	\\
	subordinates 	& 	& 	& 	\\
\hline
	Determiners 	 
	&
	\dep{det}
	& 
	\multirow{2}{*}{
	\begin{dependency}[depstd]
		\begin{deptext}[column sep=0.03cm]
		the \& book  \\
		\end{deptext}
	    \depedge[edge start x offset=3pt]{2}{1}{}
	\end{dependency}
	}
	&
	\multirow{2}{*}{
	\begin{dependency}[depstd]
		\begin{deptext}[column sep=0.03cm]
		the \& book  \\
		\end{deptext}
	    \depedge[edge start x offset=-4pt]{1}{2}{}
	\end{dependency} 
	}
	\\
	 	& 	& 	& 	\\
\hline
	Noun  
	&
	\dep{mwe}+\dep{goeswith}, 
	& 
	\multirow{2}{*}{
	\begin{dependency}[depstd,arc angle=20]
		\begin{deptext}[column sep=0.03cm]
		John \& Jr. \& Doe  \\
		\end{deptext}
	    \depedge[edge start x offset=-2pt]{1}{2}{}
        \depedge[edge start x offset=-2pt]{1}{3}{}
	\end{dependency}
	}
	&
	\multirow{2}{*}{
	\begin{dependency}[depstd]
		\begin{deptext}[column sep=0.03cm]
		John \& Jr. \& Doe  \\
		\end{deptext}
	    \depedge[edge start x offset=-2pt]{1}{2}{}
        \depedge[edge start x offset=-2pt]{2}{3}{}
	\end{dependency} 
	}
	\\
	sequences & \dep{name} & & \\
\hline
	Case  
	&
	\dep{case}
	& 
	\multirow{2}{*}{
	\begin{dependency}[depstd]
		\begin{deptext}[column sep=0.03cm]
		of \& Earth  \\
		\end{deptext}
	    \depedge[edge start x offset=4pt]{2}{1}{}
	\end{dependency}
	}
	&
	\multirow{2}{*}{
	\begin{dependency}[depstd]
		\begin{deptext}[column sep=0.03cm]
		of \& Earth  \\
		\end{deptext}
	    \depedge[edge start x offset=-3pt]{1}{2}{}
	\end{dependency} 
	}
	\\
	marking & & & \\
\hline
	Coordinations 
	&
	\dep{cc}+\dep{conj} 
	& 
	\multirow{2}{*}{
	\begin{dependency}[depstd,arc angle=20]
		\begin{deptext}[column sep=0.03cm]
		me \& and \& you  \\
		\end{deptext}
	    \depedge[edge start x offset=-3pt]{1}{2}{}
	    \depedge[edge start x offset=-3pt]{1}{3}{}
	\end{dependency} 
	}
	&
	\multirow{2}{*}{
	\begin{dependency}[depstd]
		\begin{deptext}[column sep=0.03cm]
		me \& and \& you  \\
		\end{deptext}
	    \depedge[edge start x offset=2pt]{2}{1}{}
	    \depedge[edge start x offset=-2pt]{2}{3}{}
	\end{dependency} 
	}
	\\
	~ & & & \\
\hline
	Copulas
	&
	\dep{cop}+\dep{auxpass}
	& 
	\multirow{2}{*}{
	\begin{dependency}[depstd,arc angle=30]
		\begin{deptext}[column sep=0.03cm]
		is \& nice  \\
		\end{deptext}
	    \depedge[edge start x offset=2pt]{2}{1}{}
	\end{dependency} 
	}
	&
	\multirow{2}{*}{
	\begin{dependency}[depstd]
		\begin{deptext}[column sep=0.03cm]
		is \& nice  \\
		\end{deptext}
	    \depedge[edge start x offset=-2pt]{1}{2}{}
	\end{dependency} 
	}
	\\
	~ & & & \\
\hline
\hline
	Verb
	&
	\dep{root}+\dep{aux} 
	& 
	\multirow{2}{*}{
	\begin{dependency}[depstd,arc angle=20]
		\begin{deptext}[column sep=0.03cm]
		have \& been \& done  \\
		\end{deptext}
	    \depedge[edge start x offset=3pt]{3}{2}{}
	    \depedge[edge start x offset=3pt]{3}{1}{}
	\end{dependency} 
	}
	&
	\multirow{2}{*}{
	\begin{dependency}[depstd]
		\begin{deptext}[column sep=0.03cm]
		have \& been \& done  \\
		\end{deptext}
	    \depedge[edge start x offset=-2pt]{1}{2}{}
	    \depedge[edge start x offset=-2pt]{2}{3}{}
	\end{dependency} 	
	}
	\\
	groups & & & \\
\hline
\end{tabular}
\end{small}
\caption{Annotation scheme in the UD treebanks and standard alternatives.} \label{tab:udscheme}
\end{table*}

\section{Conversion \label{sec:transfo}}

We consider several alternatives to the UD annotation scheme. Most
have been proposed by \cite{schwartz2012learnability} or have been
discussed when defining annotations of the UD (e.g. when abandoning
the so-called ``standard'' scheme of the UDT for the content-head
scheme now used in the UD). The transformations are summarized in the
upper part of Table~\ref{tab:udscheme}. We omit the transformation of
verb groups that is already analyzed in detail in
\cite{delhoneux2016representation}.  In contrast to most works
analyzing the impact of annotation conventions, the alternative
representations we consider are defined by selecting dependencies
according to their label and transforming them rather than by
modifying the tree-to-dependency conversion scheme. It is therefore
possible to apply them to any language of the UD initiative.

\subsection{From Simple Conversions...}

The syntactic relations that we transform are mostly represented with
only one dependency which can be identified by its label.  In this
case the conversion simply consists in inverting the role of the
tokens involved in the main dependency representing the syntactic
relation: the dependent becomes the head and the head becomes the
dependent.  Given an original dependency \depright{$w_i$}{$w_j$} in
which $w_i$ is the head (i.e.\ $w_i$ receive a dependency from another
word $w_h$): \textit{i)} the dependency is replaced by
\depleft{$w_i$}{$w_j$}, \textit{ii)} the former head of $w_i$, named
$w_h$, become the new head of $w_j$.  These transformations applies to
relations such as the clause subordinates (\dep{mark}), the
determiners (\dep{det}) or the case markings (\dep{case}).
  
\subsection{...to Non-Projectivity...}

\begin{figure}[ht]
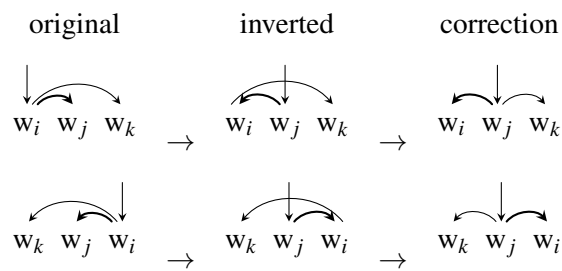

\centering
\setlength{\tabcolsep}{2.5pt}
\begin{tabular}{ccccc}
original & & inverted & & correction \\
	\begin{dependency}[depstd, edge horizontal padding=2pt]
		\begin{deptext}[column sep=0.03cm]
		w$_i$ \& w$_j$ \& w$_k$  \\
		\end{deptext}
        \deproot[edge unit distance=1.1ex]{1}{}
	    \depedge[edge start x offset=2pt, edge style={thick}]{1}{2}{}
        \depedge[edge start x offset=0pt]{1}{3}{}
	\end{dependency}
& $\rightarrow$ & 
	\begin{dependency}[depstd, edge horizontal padding=2pt]
		\begin{deptext}[column sep=0.03cm]
		w$_i$ \& w$_j$ \& w$_k$  \\
		\end{deptext}
        \deproot[edge unit distance=1.1ex]{2}{}
        \depedge[edge start x offset=0pt, edge style={thick}]{2}{1}{}
        \depedge[edge start x offset=-5pt]{1}{3}{}
	\end{dependency}
& $\rightarrow$ & 
	\begin{dependency}[depstd, edge horizontal padding=2pt]
		\begin{deptext}[column sep=0.03cm]
		w$_i$ \& w$_j$ \& w$_k$  \\
		\end{deptext}
        \deproot[edge unit distance=1.1ex]{2}{}
        \depedge[edge start x offset=0pt, edge style={thick}]{2}{1}{}
        \depedge[edge start x offset=0pt]{2}{3}{}
	\end{dependency}
\\
	\begin{dependency}[depstd, edge horizontal padding=2pt]
		\begin{deptext}[column sep=0.03cm]
		w$_k$ \& w$_j$ \& w$_i$  \\
		\end{deptext}
        \deproot[edge unit distance=1.1ex]{3}{}
	    \depedge[edge start x offset=-2pt, edge style={thick}]{3}{2}{}
        \depedge[edge start x offset=0pt]{3}{1}{}
	\end{dependency}
& $\rightarrow$ & 
	\begin{dependency}[depstd, edge horizontal padding=2pt]
		\begin{deptext}[column sep=0.03cm]
		w$_k$ \& w$_j$ \& w$_i$  \\
		\end{deptext}
        \deproot[edge unit distance=1.1ex]{2}{}
	    \depedge[edge start x offset=0pt, edge style={thick}]{2}{3}{}
        \depedge[edge start x offset=5pt]{3}{1}{}
	\end{dependency}
& $\rightarrow$ & 
	\begin{dependency}[depstd, edge horizontal padding=2pt]
		\begin{deptext}[column sep=0.03cm]
		w$_k$ \& w$_j$ \& w$_i$  \\
		\end{deptext}
        \deproot[edge unit distance=1.1ex]{2}{}
	    \depedge[edge start x offset=0pt, edge style={thick}]{2}{3}{}
        \depedge[edge start x offset=0pt]{2}{1}{}
	\end{dependency}\\
\end{tabular}
\caption{Cases of non-projectivity caused by conversion, and correction. The main (bold) dependency \depright{$w_i$}{$w_j$} is the one to invert. When inverting, $w_j$ becomes the root of the sub-structure.} \label{fig:crossingdep}
\end{figure}

  However, more than two tokens are frequently involved
in the sub-structure carried by the dependency in question. In that
case, the conversion may create non-projective dependencies (i.e.\
crossing between dependencies).  Figure \ref{fig:crossingdep}
illustrates this problem.  Let \depright{$w_i$}{$w_j$} be the original
dependency we want to invert, $w_i$ being the head and $w_j$ the dependent. If
the head $w_i$ has a child $w_k$, i.e.\ there is a $w_k$ such as \depright{$w_i$}{$w_k$}, and 
the tokens are ordered such as k$<$j$<$i or i$<$j$<$k then a crossing 
between the dependencies\footnote{A crossing generally appears between the dependency going from $w_i$ to his child $w_k$ and the root dependency, now arriving on $w_j$, i.e. coming from the former head of $w_i$.} will appear when inverting the role of $w_i$ and 
$w_j$. To avoid introducing a non-projectivity, it is
necessary to attach the former child $w_k$ of $w_i$ to $w_j$.

\def\arraystretch{0.6}

\subsection{... and Particular Cases}

\begin{figure}[ht]
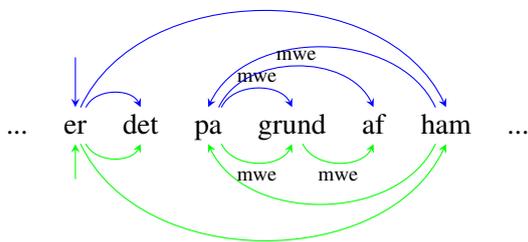

\begin{center}
\begin{dependency}[depstd,arc angle=65]
		\begin{deptext}[column sep=0.3cm]
		... \& er \& det \& pa \& grund \& af \& ham \& ...   \\
		\end{deptext}
	    \depedge[blue, edge start x offset=0pt]{2}{3}{}
	    \depedge[blue, edge start x offset=-2pt]{2}{7}{}
	    \depedge[blue, edge start x offset=0pt]{7}{4}{}
	    \depedge[blue, edge start x offset=1pt]{4}{5}{mwe}
	    \depedge[blue, edge start x offset=0pt]{4}{6}{mwe}
        \deproot[blue, edge unit distance=1.3ex]{2}{}
        
	    \depedge[green, edge start x offset=0pt, edge below]{2}{3}{}
	    \depedge[green, edge start x offset=-2pt, edge below]{2}{7}{}
	    \depedge[green, edge start x offset=0pt, edge below]{7}{4}{}
	    \depedge[green, edge start x offset=1pt, edge below, label style={below}]{4}{5}{mwe}
	    \depedge[green, edge start x offset=0pt, edge below, label style={below}]{5}{6}{mwe}
        \deproot[green, edge unit distance=1.5ex, edge below]{2}{}
\end{dependency}
\caption{Multi-word expression conversion for the danish phrase `it is because
  of him'. The dependencies following the UD conventions are
  represented in blue above the words; the alternative structure is
  represented in green below the words.}
\end{center} 
\label{fig:sequence-example}
\end{figure}

\paragraph{Noun Sequences}
For noun sequences (\dep{mwe}, \dep{name} and \dep{goeswith}), we
systematically consider the first word of the sequence as the head,
and, when the sequence contains several words, attach each word to its
preceding word, while, in UD guidelines, noun sequences are annotated
in a flat, head-initial structure, in which all words in the name
modify the first one (see Figure~\ref{fig:sequence-example}).

\paragraph{Copulas}
In copula constructions (\dep{cop} and \dep{auxpass}
  dependencies), the head of the dependency is generally the root of
  the sentence (or of a subordinate clause). The transformation of a
  copula dependency \depright{$w_i$}{$w_j$} between the the i-th and
  j-th word of the sentence consists in inverting the dependency (as
  for \dep{mark} and \dep{case}), making $w_j$ the root of the
  sentence and attaching all words that were modifying $w_i$ to~$w_j$
  with a dependency not related to nouns such as \dep{det},
  \dep{amod}, or \dep{nmod}.  The last step allows us to ensure the
  coherence of the annotations (with respect, for instance, to the
  final punctuation).

\begin{figure}[ht]
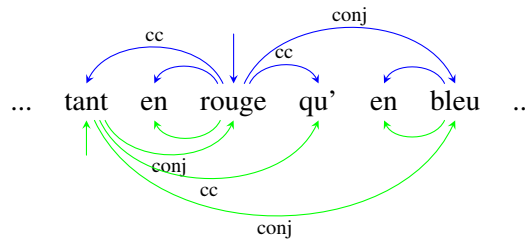

\begin{center}
\begin{dependency}[depstd,arc angle=65]
		\begin{deptext}[column sep=0.25cm]
		... \& tant \& en \& rouge \& qu' \& en \& bleu \& ...   \\
		\end{deptext}
	    \depedge[blue, edge start x offset=0pt]{4}{2}{cc}
	    \depedge[blue, edge start x offset=-2pt]{4}{3}{}
	    \depedge[blue, edge start x offset=2pt]{4}{5}{cc}
	    \depedge[blue, edge start x offset=0pt]{7}{6}{}
	    \depedge[blue, edge start x offset=0pt]{4}{7}{conj}
        \deproot[blue, edge unit distance=1.3ex]{4}{}
        
	    \depedge[green, edge start x offset=3pt, edge below, label style={below}]{2}{4}{conj}
	    \depedge[green, edge start x offset=-1pt, edge below]{4}{3}{}
	    \depedge[green, edge start x offset=1pt, edge below, label style={below}]{2}{5}{cc}
	    \depedge[green, edge start x offset=0pt, edge below]{7}{6}{}
	    \depedge[green, edge start x offset=-1pt, edge below, label style={below}]{2}{7}{conj}
        \deproot[green, edge unit distance=1.5ex, edge below]{2}{}
\end{dependency}
\end{center}
\caption{Coordination conversion for the French phrase `as well in red
  as in blue'. The dependencies following the UD conventions are
  represented in blue above the words; the alternative structure is
  represented in green below the words.} \label{fig:coord-example}
\end{figure}

\paragraph{Coordinations}
For coordinating structures (\dep{cc} and \dep{conj}
dependencies), in the UD scheme, the first conjunct\footnote{Typically
  a noun for instance (but could also be a verb or an adjective) for
  which the incoming dependency could be labeled with \dep{dobj},
  \dep{root}, \dep{amod}, etc.} is taken as the head of the
coordination and all the other conjuncts depend on it via the
\dep{conj} relation, and each coordinating conjunction\footnote{Often
  PoS-tagged with a \textsc{CONJ} such as \emph{and}, \emph{or},
  etc.} is attached to the first conjunct with a \dep{cc}
relation.\footnote{Recall that we are considering the version 1
  guidelines; the definition of the \dep{cc} relation has changed in
  version~2.} As an alternative, we define the first coordinating
conjunction as the head and attach all conjuncts to it (see Figure
\ref{fig:coord-example}).

\section{Experimental Settings} \label{sec:experiments}

To evaluate the proposed transformations, we follow the approach
introduced in \cite{schwartz2012learnability} consisting in
comparing the original and the transformed data on their respective
references. 

\subsection{Data}

We experiment on data from the v1.3 of the Universal Dependency
project \cite{universaldependencies13}, using the official split into
train, validation and test sets. We apply separately the 7 transformations
described in Section~\ref{sec:transfo} on the 38 languages of the UD,
resulting in the creation of 266 transformed corpora, 44 of which were
identical to the original corpora as the transformation can not be
applied (e.g.\ there are no multi-word expressions in Chinese). These
corpora are not included in the different statistics presented in this
Section.

For each configuration (i.e.\ a language and a transformation), 
a dependency parser is trained on the original
data annotated with UD convention (denoted \texttt{UD}) and the transformed data (denoted
\texttt{transformed}). Parsing performance is estimated using the
usual Unlabeled Attachment Score (UAS, excluding punctuation). 
Reported scores are averaged over three trainings.

\subsection{Parser}

We use our own implementation of the arc-eager dependency parser with
a dynamic oracle and an averaged
perceptron~\cite{Aufrant:2016:PanParser}, using the features described
in~\cite{zhang11transition} which have been designed for
English. 
Preliminary experiments show that similar results are achieved with
other implementation of transition-based parsers (namely with the
MaltParser \cite{Nivre03efficient}).

\section{Results\label{sec:results}}

\begin{figure}[h]
  \includegraphics[width=\columnwidth]{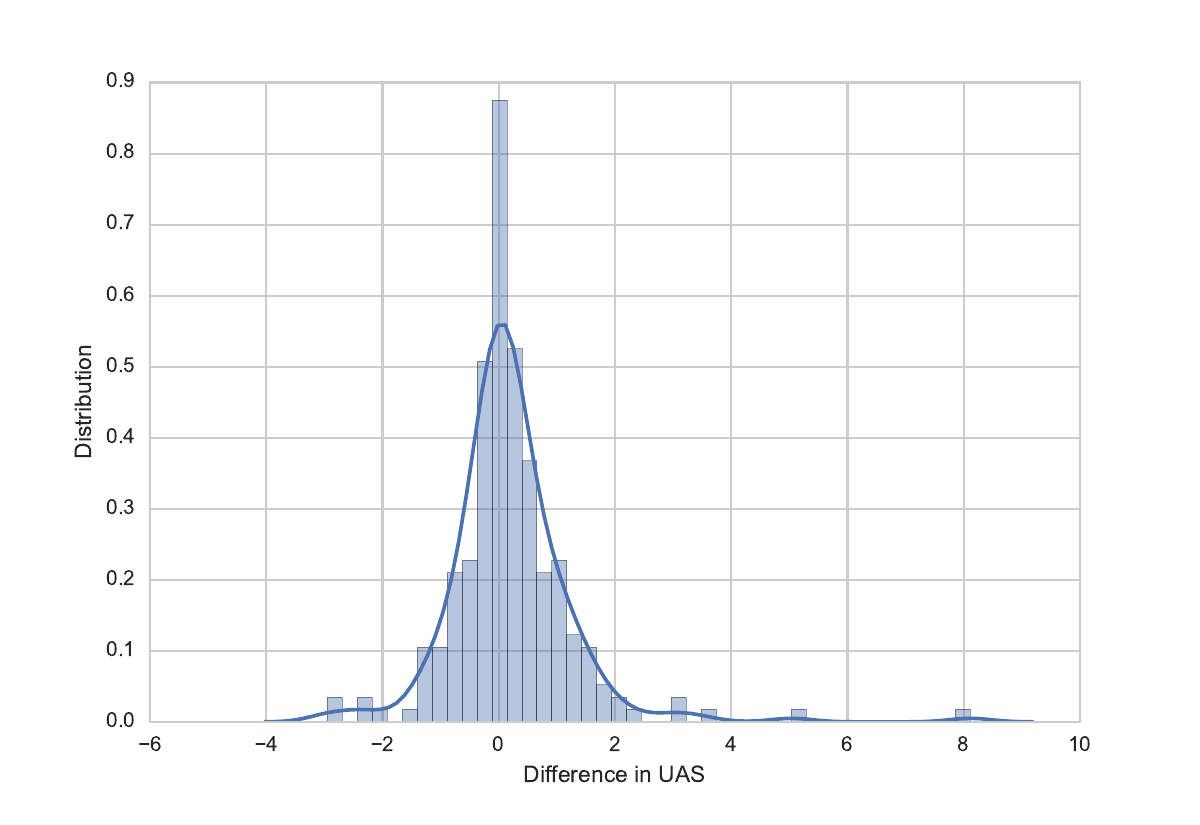}
  \vspace{-2em}
  \caption{Distribution of differences between the UAS achieved on the
    UD and transformed corpora for the different languages and
    transformations considered. Positive differences indicates better results with UD annotations. 
    \label{fig:distrib_uas_diff}}
\end{figure}

Figure~\ref{fig:distrib_uas_diff} shows the distribution of
differences in UAS between a parser trained on the original data and a
parser trained on the transformed data (positive differences indicates
corpora for which the UD annotation scheme results in better
predictions). As expected, the annotation scheme has
a large impact on the quality of the prediction, with an average
difference in scores of 0.66 UAS points and variations as large as
8.1~UAS points.

However, contrary to what is usually believed, the UD scheme appears
to achieve, in most cases, better prediction performance than the
proposed transformations: in 58.1\% of the configurations, the parser
trained and evaluated on transformed data is outperformed by the
parser trained on the original UD data. More precisely, the difference
in UAS is negative in 93 configurations and positive in 129
configurations. Table~\ref{tab:uas_diff} details for each
transformation the percentage of languages for which the UD scheme
results in better predictions. The \dep{cc} dependency (conjunction),
and to a lesser extent the \dep{det} dependency, are easier to learn
in the UD scheme than in the proposed transformed scheme. On the
contrary, the choice of the \dep{cop} and \dep{name} structure in the
UD results in large losses for many languages. For the other
variations considered, the learnability of the scheme highly depends
on the language.  Table~\ref{tab:diff} shows the configurations with
the largest positive and negative differences in scores.

\def\arraystretch{0.7}
\begin{table}[h]
  \begin{tabular}{@{~}l@{~~}c@{~~~~~}l@{~~}c@{~~~~~}l@{~~}c@{~}}
    \toprule
    \dep{case}        &  44.74\% & \dep{mark}        &  58.33\% & \dep{det}         &  80.56\% \\
    \dep{cc}          &  89.47\% & \dep{mwe}         &  50.00\% & \dep{name}        &  45.83\% \\
    \dep{cop}         &  25.00\% \\
    \bottomrule
  \end{tabular}
  \centering
  \caption{Number of times, for each transformation, a parser trained
    and evaluated on UD data outperforms a parser trained and
    evaluated on transformed data.\label{tab:uas_diff}}
\end{table}

\begin{table}[h]
  \centering
  \begin{tabular}{llrr}
    \toprule
    Lang. & Transfo. & UAS(trans.) &  UAS(UD) \\
    \midrule
    la &  conjunction &      52.57\% & \textbf{60.69\%} \\
    gl &         case &      72.17\% & \textbf{77.21\%} \\
    ar &  conjunction &      72.35\% & \textbf{75.83\%} \\
    kk &         case &      56.62\% & \textbf{59.67\%} \\
    zh &         mark &      66.67\% & \textbf{69.65\%} \\
    \midrule
    nl &       copule &      \textbf{69.82\%} & 67.73\% \\
    fi &       copule &      \textbf{66.59\%} & 64.30\% \\
    et &       copule &      \textbf{70.38\%} & 67.95\% \\
    la &       copule &      \textbf{59.34\%} & 56.47\% \\
    sl &       copule &      \textbf{79.69\%} & 76.75\% \\
    \bottomrule
  \end{tabular}
  \caption{Languages and transformations with the highest UAS
    difference.~\label{tab:diff}}
\end{table}

\begin{table}[h]
  \centering
  \begin{tabular}{lc}
    \toprule
    metric                & \\
    \midrule
    distance              & 43.6\% \\
    predictability        & 64.8\% \\
    derivation complexity & 62.6\% \\
    derivation perplexity & 61.2\% \\
    \bottomrule
  \end{tabular}
  \caption{Number of times a given learnability measure is able to
    predict which annotation scheme will result in the best parsing
    performance. \label{tab:learnability}}
\end{table}

\paragraph*{Analysis}

To understand the empirical preferences of annotation schemes we
consider several measures of the `learnability' and `complexity' of a
treebank:
\begin{itemize}
\item the average absolute \emph{distance} (in words) between a
dependent and its head;
because 
transition-based dependency parsers are known to favor short
dependencies over long
ones~\cite{mcdonald07characterizing};
\item the \emph{predictability} of the scheme introduced by
\cite{schwartz2012learnability} defined as the entropy of the
conditional distribution of the PoS of the dependent knowing the PoS
of its head;
\item the \emph{derivation perplexity} introduced by
  \cite{sogaard10derivation} defined as the perplexity of 3-gram
  language model estimated on a corpus in which words of a sentence
  appear in the order in which they are attached to their
  head;\footnote{Similarly to~\cite{sogaard10derivation} we consider a
    trigram language model but use a Witten-Bell smoothing as many
    corpora were too small to use a Knesser-Ney smoothing.  As for the
    derivation complexity, the words are ordered according to an
    oracle prediction of the reference structure.}
\item the \emph{derivation complexity} defined as the sum of the
number of distinct substrings in the gold derivations of the corpora
references.
\end{itemize}
The first three metrics have been used in several studies on the
learnability of dependencies annotations. We introduce the last one as
a new way to characterize the difficulty of predicting a sequence of
actions, building on the intuition that the more diverse a derivation,
the harder its prediction. For this
metric, 
the gold derivation is the concatenation of arc-eager actions
representing the sequence of actions generating a reference tree. In
case of ambiguity in the generation of the reference tree, we always
select the actions in the following order: \textsc{Shift},
\textsc{Reduce}, \textsc{Left}, \textsc{Right}. Using a generalized
suffix tree it is then possible to count the number of different
substrings in the derivations with a complexity in
$\mathcal{O}(n)$~\cite{gusfield97algorithms}.

A metric is said \emph{coherent} if it scores the syntactic
structure that achieves the best parsing performance higher than its
variation. Table~\ref{tab:learnability} reports the numbers of times, averaged
over languages and transformations, that each metric is
coherent.

Contrarily to what has been previously reported, the considered
metrics are hardly able to predict which annotation scheme will result
in the best parsing performance. Several reasons can explain this
result. First, it is the first time, to the best of our knowledge that
these metrics are compared on such a wide array of languages. It is
possible that these metrics are not as language-independent as can be
expected. Second, as our transformations are directly applied on the
dependency structures rather than when converting the dependency
structure from a constituency structure, it is possible that some of
their transformations are erroneous and the resulting complexity
metric biased.

\section{Conclusion}

Comparing the performance of parsers trained and evaluated on UD data
and transformed data, it appears that the UD scheme leads mainly to
better scores and that measures of learnability and complexity are not
sufficient to explain the annotation preferences of dependency
parsers.

\section{Acknowledgement}

Oph\'{e}lie Lacroix is funded by the ERC Starting Grant LOWLANDS No. 313695.

\bibliographystyle{acl}
\bibliography{nodalida2017}



\end{document}